\documentclass[12pt]{custarticle}

\usepackage{url}

\usepackage{breakurl}
\usepackage[hidelinks,breaklinks]{hyperref}
\usepackage{float}
\usepackage{verbatim} 
\usepackage{apalike}
\usepackage{mathtools} 
\usepackage{booktabs} 
\usepackage{tikz} 
\usepackage{color}
\usepackage{bbm}
\usepackage[margin=1in]{geometry}
\usepackage{setspace}
\usepackage{graphicx}	
\usepackage{algorithm}
\usepackage{multicol}
\usepackage{amsthm}
\usepackage{amssymb}
\usepackage{mathtools}
\usepackage{subcaption}
\usepackage{makecell}
\usepackage{url}

\makeatletter
\def\ps@pprintTitle{%
  \let\@oddhead\@empty
  \let\@evenhead\@empty
  \let\@oddfoot\@empty
  \let\@evenfoot\@oddfoot
}
\makeatother

\biboptions{authoryear}

\begin{document}
\begin{frontmatter}

\title{User Friendly and Adaptable Discriminative AI: Using the Lessons from the Success of LLMs and Image Generation Models}

\author[label1]{Son The Nguyen}
\ead{snguye65@uic.edu}

\author[label1]{Theja Tulabandhula}
\ead{theja@uic.edu}

\author[label1]{Mary Beth Watson-Manheim}
\ead{mbwm@uic.edu}

\address[label1]{University of Illinois Chicago, 601 S Morgan St, Chicago, IL 60607, United States}

\begin{abstract}
While there is significant interest in using generative AI tools as general-purpose models for specific ML applications, discriminative models are much more widely deployed currently. One of the key shortcomings of these discriminative AI tools that have been already deployed is that they are not adaptable and user-friendly compared to generative AI tools (e.g., GPT4, Stable Diffusion, Bard, etc.), where a non-expert user can iteratively refine model inputs and give real-time feedback that can be accounted for immediately, allowing users to build trust from the start. Inspired by this emerging collaborative workflow, we develop a new system architecture that enables users to work with discriminative models (such as for object detection, sentiment classification, etc.) in a fashion similar to generative AI tools, where they can easily provide immediate feedback as well as adapt the deployed models as desired. Our approach has implications on improving trust, user-friendliness, and adaptability of these versatile but traditional prediction models.
\end{abstract}


\end{frontmatter}

\section{Introduction}\label{sec:intro}

Machine learning (ML) methods, a subset of the wider artificial intelligence (AI) field, can be divided into discriminative and generative approaches. Discriminative methods focus on modeling the conditional probability of outcome(s) given a context (such as a feature vector). In contrast, generative methods focus on modeling the joint distribution of data. Discriminative models have historically found success in classification and regression tasks in various domains (e.g., finance, healthcare, automotive, etc). On the other hand, newer generative models, such as Large Language Models (LLMs) and diffusion models, have succeeded in open-ended tasks that require versatility and creativity in addition to traditional prediction tasks. We hypothesize that the value of these new generative models is enhanced because they are user-friendly and highly adaptable, making it easier for non-experts to interact with them and produce valuable results with minimal effort. However, this is not the case with current discriminative models. In this work, we explore ways to make discriminative models more user-friendly and adaptable, which we hypothesize will increase their adoption in more applications and bring them on par with the success levels seen with generative AI tools.

Modern generative models have billions of parameters and are trained on vast amounts of publicly available data on the internet, making them highly complex and powerful \citep{shanahan2023talking}. They are capable of generating coherent and innovative content, as well as excelling in discriminative tasks such as classification, sentiment analysis, and named entity recognition. The interactive paradigm they offer is akin to having a dialogue with a multitool, allowing users to guide and collaborate with AI in an intuitive and responsive manner through prompting. This \emph{transformative interaction} enables a sense of control and has opened avenues for unprecedented advancements, allowing generative models to address a myriad of applications and challenges. Innovations like OpenAI’s DALL-E, which is capable of crafting coherent images from textual descriptions, and ChatGPT, which is designed to enrich human-AI interactions, serve as prime exemplars. 

Discriminative models, on the other hand, have not been as widely adopted as generative models because users tend to avoid adopting black-box algorithms that: (a) only provide a response or decision without any explanation \citep{SAEED2023110273}, and/or (b) don't appear to or allow for very limited real-time feedback and collaboration \cite{AlgorithmicRelationships}. Based on the success of generative AI tools, we propose a more \emph{agile and responsive discriminative AI system architecture}. This architecture allows for real-time feedback from end-users and real-time improvement of the underlying discriminative models, giving them direct control over these systems. In addition to increasing the user's sense of control, these changes can enable long-term trust and companionship. Thus, one of our goals is to create interfaces and systems that enable users to interact dynamically with the AI tool and influence it as their needs change over time. As shown in Section~\ref{sec:improveddiscai}, our proposed architecture emphasizes real-time adaptability of the underlying ML models, ensuring optimal performance in real-time scenarios as well as user-friendliness. We posit that such systems architectures can amplify the adoption of discriminative models similar to the success we have seen with generative models.

Modern generative models are now able to perform better during inference with less data than before by using pre-trained foundation models. These models can be adapted to few-shot \citep{DBLP:journals/corr/abs-2005-14165} or even zero-shot learning \citep{DBLP:journals/corr/abs-2109-01652} with ease without fine-tuning, thus casting doubt on whether improving the user experience when working with discriminative models is even useful. It has been observed that, despite the power of modern (multimodal) generative models, utilizing them for all tasks is not suitable for multiple reasons. Firstly, these generative models currently use natural language interfaces, which we know are inefficient for all tasks (e.g., real-time vision tasks). Second, these models effectively sample new data points (via their responses, generated images, etc.) which may not be grounded in reality \citep{zhang2023sirens} or sufficiently constrained \citep{wolf2023fundamental}. Third, these models may require much more computation and data than an equivalent discriminative model built for the prediction task at hand, making it impractical for small developers to train from scratch and operate. This creates a dependence on large AI enterprises and their foundation models, raising concerns about security and privacy issues \citep{nasr2023scalable}.

Therefore, discriminatory models continue to be a compelling alternative for many situations. However, the inflexibility of discriminative AIs presents a significant challenge in application, limiting users' ability to modify the model's behavior to fit and adapt to their specific needs. For example, discriminative models enabling autonomous driving are generally not improvable in real-time today, and the users (drivers) do not have direct access to them. This lack of adaptability and user-friendliness has led to many situations where failures could be avoided if these models were more user-friendly and adaptable. For instance, activists have disabled autonomous vehicles by placing traffic cones on their hoods, taking advantage of the models' detection patterns \citep{Curry_2023}. In addition, there have been cases where Tesla's Autopilot system has been tricked into speeding \citep{2020} or failed to detect responding vehicles, leading to accidents \citep{autoevolution_2023}. These incidents highlight the need for AI solutions that go beyond technical proficiency. We posit that, it is important to enhance user-friendliness and adaptability of discriminative models by incorporating features that allow the user to provide feedback, make quick corrections, and enable personalization of the models. By doing so, these models can improve their performance and build immediate trust with users, ultimately accelerating their adoption at large. In the above example related to autonomous driving, if the drivers are given the ability to provide feedback, fix errors, and personalize their models quickly, they may help improve the models using data from the cases where the models fail. The models are then personalized better to their local environments and can detect objects more accurately, which could mitigate catastrophic and chain accidents and improve overall driving safety and experience.

\begin{figure}[h]
    \centering
    \subcaptionbox{Activists disabling autonomous vehicles by placing bright orange traffic cones on their hoods \citep{Curry_2023}.}{\includegraphics[height=3cm]{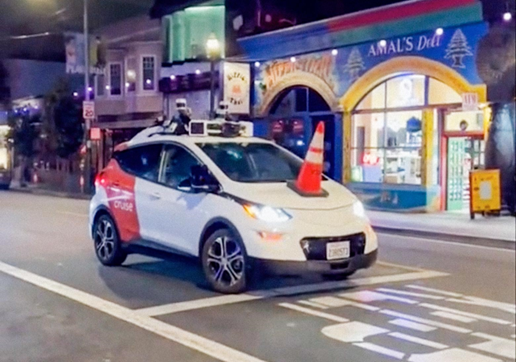}}\hfill
    \subcaptionbox{Tesla Being Tricked into Speeding \citep{2020}.}{\includegraphics[height=3cm]{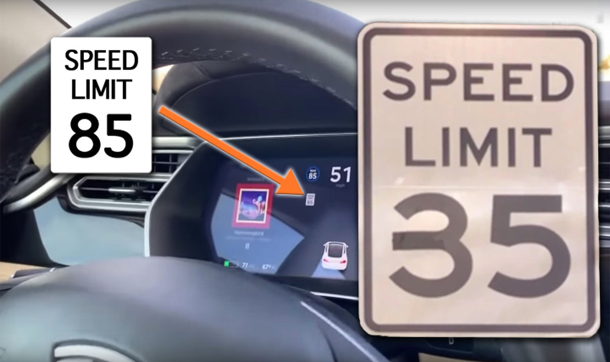}}\hfill
    \subcaptionbox{Tesla Autopilot failed to detect responder vehicles, resulting in accident \citep{autoevolution_2023}.}{\includegraphics[height=3cm]{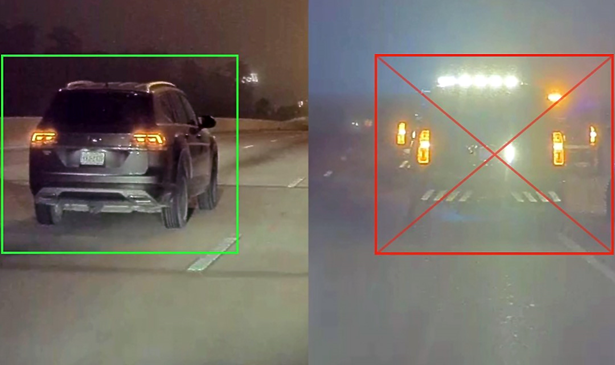}}
    \caption{Examples that illustrate the need for continuous feedback/adaptation of the underlying model based on feedback.}
    \label{fig:incidentexamples}
\end{figure} 

Our key contribution is the following. We propose a discriminative ML system architecture that allows end-users, rather than ML experts, to handle data-related issues and improve models that have already been deployed. The system is designed to have the ability to constantly adapt and correct itself without any downtime by incorporating user feedback that uses the same language as the model output. Our system architecture is inspired by the iterative interaction loops between users and models observed in generative AI systems, to improve the adaptability and user-friendliness of discriminative models.

\section{Related Work}\label{sec:relatedwork}
\subsection{Algorithmic Aversion and Automation Bias}\label{subsec:aversionauto}

Research identifies two contrasting attitudes towards AI that emerge during user interactions with such systems: algorithmic aversion and automation bias.

Algorithm aversion occurs when individuals disregard reliable algorithmic predictions and prefer less accurate user judgments \citep{Dietvorst2015}. Some studies suggest that users avoid using algorithms due to their intolerance towards errors \citep{Dietvorst2015, Highhouse}, their lack a sense of control over the decisions at hand \citep{doi:10.1287/mnsc.2016.2643, Cheng_Chouldechova_2023, 10.1093/jcmc/zmac029, SchaapABC}, or the uninterpretability of the predictions.\citep{ethnographer,Lebovitz2021IsAG,LebovitzOpacity}

Automation bias, on the other hand, is the tendency of users to overly rely on automated suggestions without critical evaluation, leading to decision-making errors \citep{mosier1996automation,cummings2004automation}. This bias can occur when users rely on heuristics, perceive automated decision-making tools as superior, or reduce their effort when working redundantly within a group \citep{doi:10.1177/0018720810376055}. While research on the effectiveness of using explainable AI to combat this problem has produced mixed results, most current studies recommend its use as a solution \citep{VERED2023103952, vasconcelos2023explanations, schemmer2022influence}.

\subsection{Contemporaenous work on improving discriminative AI systems}\label{subsec:discriminativeai}
Efforts are being made to improve discriminative AI systems to make them beneficial to users. \cite{10.1145/3411763.3451798} enhances human-AI interaction by both explaining AI decisions to users and interpreting user responses to improve AI performance. \cite{Miao_2021} proposes a human-in-the-loop framework where human experts and AI mutually enhance each other via model updates and human intervention. \cite{noti2023learning} designs a system architecture incorporating a learning advising policy to determine if AI predictions should be displayed to users.

The aforementioned studies have explored the communication between discriminative AI systems and users. However, they follow a traditional system architecture where the fine-tuning step is separate from the inference step, which means that users cannot control model updates, and updates do not occur immediately. In contrast, our proposed machine learning architecture is designed to be user-friendly, allowing end-users to control the fine-tuning process by providing feedback on data-related issues. Also, the system is equipped with an iterative fine-tuning process to learn and adapt to user feedback quickly.

\section{Preliminaries}\label{sec:preliminaries}

We now present some key properties that useful AI systems should possess and evaluate how current discriminative AI and generative AI system architectures perform in terms of these properties.

\subsection{The Five Attributes of Useful AI} \label{subsec:useful-ai}

Similar to how humans are evaluated with respect to their hard and soft skills, one can envision a similar approach to evaluating AI systems. For instance, it is critical for AI systems to have high performance and efficiency metrics, which can count as their hard skills. Additionally, their soft skills, such as versatility, user experience, and adaptability, can also be considered important for the overall success of the system or service they are a part of. Based on these ideas, we introduce \emph{Useful AI} as a fundamental concept in our discussion. Useful AI is built on five foundational pillars: performance, efficiency, versatility, user-friendliness, and adaptability. These properties are defined as follows:

\begin{itemize}
    \item \textit{Performance}: Defines the system's proficiency (e.g., error rates) and reliability in task execution.
    \item \textit{Efficiency}: Denotes the system's capability to perform tasks quickly with minimal resource consumption.
    \item \textit{Versatility}: Reflects the system's capability to operate effectively in diverse situations and tasks.
    \item \textit{User-friendliness}: Describes the system's intuitive and engaging experience through responsive and interactive functionalities. 
    \item \textit{Adaptability}: Reflects the system's capacity for timely adjustment to individual requirements and preferences.
\end{itemize}

We now analyze and contrast the generative and current discriminative AI architectures typically seen in practice based on these Useful AI properties above. Our objective is to present the advantages and disadvantages of each system and identify addressable gaps. In particular, these observed gaps lead us to develop a discriminative AI system that can enhance soft skills, an area where modern generative AI performs exceptionally. We will discuss this in detail in Section \ref{sec:improveddiscai}.

\subsection{Current Discriminative AI Architecture}\label{subsec:currentdiscriminative}

\begin{figure}
    \centering
    \includegraphics[width=0.75\linewidth]{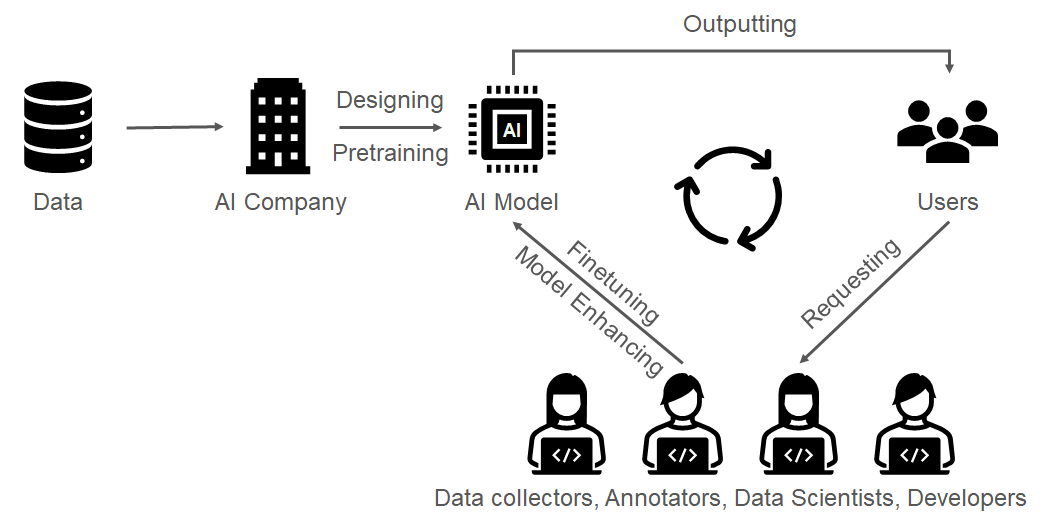}
    \caption{Baseline discriminative AI system architecture.}
    \label{fig:disai}
\end{figure}

Current discriminative AI systems follow a static development lifecycle as shown in Figure~\ref{fig:disai}. This usually involves a joint effort between ML experts and users to gather and label the required datasets. Once the data is annotated, ML experts either fine-tune an existing pre-trained neural network architecture or train a model from scratch using the data. Once trained/fine-tuned, the models remain in production with minimal changes. Although user feedback mechanisms and drift detection systems are implemented to ensure the quality of the model stays consistent over time, user feedback is not integrated immediately. Updates occur only when specific thresholds of time or performance degradation are met and are carried out under the supervision of the ML experts. 

We now discuss how the system fares with respect to the properties discussed in Subsection~\ref{subsec:useful-ai}. Regarding hard skills, we observe that discriminative models fare well in performance measures (e.g., accuracy in classification and prediction tasks) and require less training data and computation than their generative counterparts (thus being more efficient). However, in terms of soft skills, they have several weaknesses. Firstly, they lack versatility and need to be re-trained to adapt to various use cases. This limitation is inherent in discriminative AI and requires considerable effort to change \citep{DBLP:journals/corr/abs-2103-00020}. This is because there is a trade-off between efficiency and versatility. On the other hand, we posit that properties such as adaptability and user-friendliness are not inherently limiting these systems and can be fixed (see the proposed architecture in Section \ref{fig:improveddisai}).

In terms of adaptability, it has been observed in practice that discriminative models are rarely implemented to be customized in real-time. In particular, a significant drawback of these systems is the delay between the detection of data changes or model errors by the users, and the updating of the model to reflect those changes. This delay can be problematic, especially in applications where real-time accuracy is crucial. Outdated models in rapidly evolving domains can lead to catastrophic accidents, as discussed in Section \ref{sec:intro}. This, in turn, can cause users to lose their trust and confidence in the system, leading them to avoid using it and thus decreasing the system-level output.

Similarly, drawbacks also exist with respect to user-friendliness. We have observed that many deployed discriminative models are not optimized for interactive human-in-the-loop usage (i.e., offer little to no interactivity between the users and the models). An issue that exacerbates this is the current way of generating and managing labeled training data. Quality data labeling requires human-in-the-loop workflows to gather reliable labeled datasets for effective ML. While consistency and accuracy are paramount at this labeling stage, the task is often performed by a separate set of annotators rather than the end-users of the model. This delegation can introduce potential issues with the trained model, which end-users could correct if the AI system is more user-friendly. For instance, despite their best efforts, annotators may lack the specialized knowledge and insights actual end-users have when using the AI system. As a result, the labeled data might exhibit inconsistencies or miss nuanced details that only end-users would recognize and are in a position to correct. These missed details compromise the performance of the model. Even if the users are given the ability to give feedback about the models, the current interfaces often do not facilitate seamless and intuitive fixing of these errors. While there have been attempts in the literature to incorporate mechanisms that help with a sense of control, we posit that modern human-in-the-loop workflows also fundamentally require building strong trust and companionship. And one of the modalities to achieve this is via system designs that enable user-friendly usage and feedback.

\subsection{Modern Generative AIs Architecture}\label{subsec:moderngenerative}

\begin{figure}
    \centering
    \includegraphics[width=0.75\linewidth]{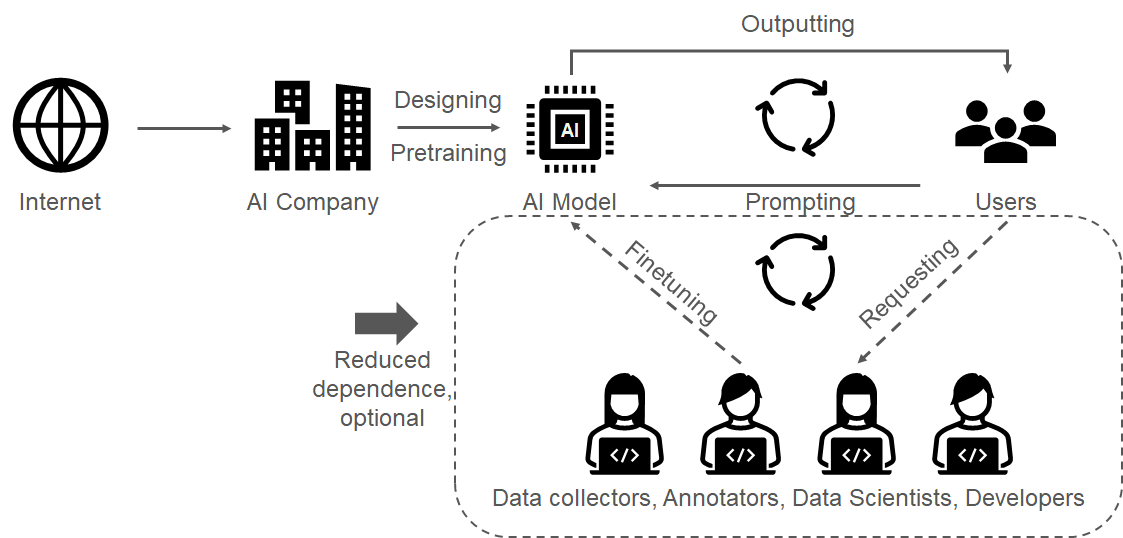}
    \caption{Modern generative AI system architecture.}
    \label{fig:genai}
\end{figure}

Generative AI systems still follow a fixed development process, as shown in Figure~\ref{fig:genai}. Large AI enterprises train large generative models with extensive datasets, equipping them with a broad knowledge base to accurately perform diverse, real-world tasks \cite{shanahan2023talking}. During inference, the model allows users to communicate with it using natural language queries, enabling conversations or modifications of its results. This creates a direct feedback mechanism between the user and the AI, fostering intuitive collaboration. Users are now focused on designing, developing, and refining prompts to effectively utilize and enhance the capabilities of large language models (LLMs) for their tasks. So generally, there's no need for regular fine-tuning or complete retraining of the model, except for specialized or complex tasks or updating its knowledge base.

We now examine how the generative AI architecture meets the five properties of Useful AI. Generative models perform well with few-shot or zero-shot prompting if they are large and trained on lots of data \cite{shanahan2023talking}. The training needs to be distributed across hundreds of GPUs (thus, their performance is great but less efficient). In terms of the soft skills we considered, they do not have any weaknesses. Firstly, they thrive in versatility and don't need to be re-trained to adapt to various use cases. This advantage is inherent in generative AI due to the trade-off between efficiency and versatility. The user-friendly nature of these systems is highlighted by their turn-by-turn natural language interfaces, which facilitate conversational interactions. This feature allows users to accomplish their goals easily without encountering difficult learning curves. Moreover, generative models can adapt in real-time to conversations using context length for memory retention \citep{liu2023lost}. Feedback loops with natural language prompts enable precise tailoring of responses. External data and APIs can also be incorporated for up-to-date knowledge retrieval \cite{DBLP:journals/corr/abs-2005-11401}.

\begin{table}[h]
\centering
\begin{tabular}{|l|c|c|c|}
\hline
\textbf{Attribute}     & \textbf{Generative AI} & \textbf{\makecell{Current \\Discriminative AI}} & \textbf{\makecell{Proposed \\Discriminative AI}} \\ \hline
Performance     &High&High&High \\ \hline
Efficiency      &Low&High&High \\ \hline
Versatility     &High&Low&Low \\ \hline
User-friendliness &High&Low&High \\ \hline
Adaptability    &High&Low&High \\ \hline
\end{tabular}
\caption{System architectures with different types of models (discriminative and generative) and how they fare with respect to Useful AI properties.}
\label{table-1}
\end{table}

\section{Proposed Discriminative AI Architecture}\label{sec:improveddiscai}

Given the limitations of the conventional approach to discriminative AI architecture, there's a growing consensus about the need for more agile system architecture methodologies that are more adaptable and user-friendly for real-time changes.

\begin{figure}
    \centering
    \includegraphics[width=0.75\linewidth]{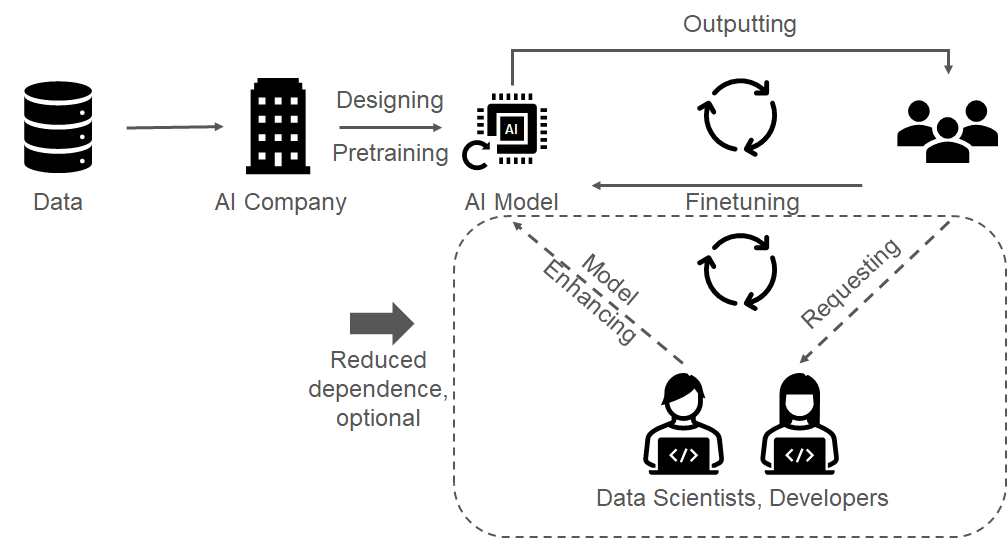}
    \caption{Proposed discriminative AI architecture that's user-friendly and adaptable.}
    \label{fig:improveddisai}
\end{figure}

The proposed discriminative AI systems follow a more agile development process, as shown in Figure~\ref{fig:improveddisai}. In this collaborative approach, roles are precisely defined within two feedback cycles: ``Developer-in-the-loop" and ``User-in-the-loop." In their cycle, developers concentrate on the model's technical refinement, addressing algorithmic challenges and integrating insights from deployment. Concurrently, users become the AI's primary instructors in their loop, collecting and curating data, flagging inaccuracies or mismatches, and ensuring the data's relevance and quality. Users have the ability to start fine-tuning with a pre-trained model that developers have prepared and incremental data during inference. The system's architecture allows users to flag mistakes or mismatches between the model's predictions and the ground truths and provide corrections through an intuitive user interface. Users can use feedback language closely aligned with the model's output format. For instance, for classification tasks, users can correct prediction labels. Similarly, for object detection in the computer vision domain, users can adjust bounding box positions and labels of objects on images. The model is then updated in real-time with feedback without waiting for a certain interval.

\textit{Adaptability Improvement:} The proposed method employs an iterative fine-tuning process to enhance adaptability compared to the baseline discriminative AI. In this approach, detection models evolve from static to dynamic detectors, continuously improving their decision-making based on consistent feedback from users. This ensures that the systems can continuously learn from errors, data shifts, edge cases, and new classes in real-time, staying updated with the most recent user insights.

\textit{User-friendliness Improvement:} The proposed method also improves user-friendliness by integrating interactive UIs. These UIs serve as a connection between users and the AI, allowing them to observe and participate in shaping the AI's predictions. We expect that this can boost user engagement and help them establish control and transparency with the AI systems. We believe that such active user involvement is essential in countering automation bias and reducing algorithm aversion.

\textit{Performance and Efficiency-Versatility Tradeoff:} The proposed discriminative model architecture maintains or even enhances its performance measures. This is because it is highly adaptable, capable of correcting errors in real-time, and allows for user collaboration. Still, it requires less data than generative models, making it more efficient. However, the trade-off between efficiency and versatility is still inherent. In summary, performance, efficiency, and versatility measures remain similar to the current discriminative AI architecture.

All these improvements converge towards a singular objective: aligning AI systems with user expectations. Leveraging real-time feedback and improvement mechanisms, these systems consistently recalibrate to match user-defined expectations. In this case, active user involvement can lead to a mutually beneficial relationship between users and AI. When users see that their feedback directly influences model training, they realize the potential of AI guided by their knowledge, and how it can enhance their task. This recognition can lead to increased alertness and attention to detail in giving feedback, as users know their careful feedback will improve the system's performance and ease their work. Their active participation and the visible impact of their expertise on the system's performance can instill a deeper sense of trust in the AI system.

\section{Conclusion}
We propose and discuss five aspects of Useful AI and identify how current discriminative AI and modern generative AI systems fare with respect to these. Based on this, we focus on the adaptability and user-friendliness properties of current discriminative AI systems and propose a new system architecture that improves on these two fronts by enabling real-time feedback and improvement.

Our approach can make discriminative AI more relevant, trustworthy, and useful to a broader range of users and provide a way to mimic the success that generative AI tools have seen in society. The collaborative partnership that the new system enables can ultimately help users meet their expectations and achieve better results.

\bibliography{reference}

\end{document}